\documentclass[sigconf,nonacm]{acmart}
\newcommand{\sys}{\textsc{IFCA-MIR}\xspace}
\usepackage{amsmath,amsfonts}
\usepackage{algorithmic}
\usepackage{graphicx}
\usepackage{subcaption}
\usepackage{textcomp}
\usepackage{xcolor}
\usepackage{physics}
\usepackage{enumitem}
\usepackage{thmtools} 
\usepackage{booktabs} 
\usepackage{multirow}
\usepackage{thm-restate}
\usepackage{bbm}
\usepackage[abbreviations]{foreign}
\theoremstyle{definition}
\usepackage{ifthen}
\usepackage{cleveref}
\theoremstyle{remark}
\newtheorem{remark}{Remark}
\newtheorem{theorem}{Theorem}

\newtheorem{corollary}[theorem]{Corollary}

\newtheorem{assumption}{Assumption}
\usepackage[algo2e,ruled,lined,boxed,commentsnumbered, noend, linesnumbered, procnumbered]{algorithm2e}

\allowdisplaybreaks

\DeclareMathOperator*{\argmin}{arg\,min}

\begin{document}

\title{Mitigating Membership Inference Vulnerability in Personalized Federated Learning}


\author{Kangsoo Jung}
\orcid{0000-0003-2070-1050}
\affiliation{%
  \institution{INRIA}
  \city{Palaiseau}
  \state{}
  \country{France}}
\email{gangsoo.zeong@inria.fr}

\author{Sayan Biswas}
\orcid{0000-0002-2115-1495}
\affiliation{%
  \institution{EPFL}
  \city{Lausanne}
  \country{Switzerland}}
\email{sayan.biswas@epfl.ch}

\author{Catuscia Palamidessi}
\orcid{0000-0003-4597-7002}
\affiliation{%
  \institution{INRIA and LIX, \'Ecole Polytechnique}
  \city{Palaiseau}
  \country{France}
}
\email{catuscia@lix.polytechnique.fr}


\renewcommand{\shortauthors}{Jung et al.}

\begin{abstract}
Federated Learning (FL) has emerged as a promising paradigm for collaborative model training without the need to share clients' personal data, thereby preserving privacy. However, the non-IID 
nature of the clients' data introduces major challenges for FL, highlighting the importance of personalized federated learning (PFL) methods. In PFL, models are trained to cater to specific feature distributions 
present in the population data. A notable method for PFL is the Iterative Federated Clustering Algorithm (IFCA), which mitigates the concerns associated with the non-IID-ness by grouping clients with similar data distributions. While it has been shown that IFCA enhances both accuracy and fairness, its strategy of dividing the population into smaller clusters 
increases vulnerability to Membership Inference Attacks (MIA), particularly among minorities with limited training samples. In this paper, we introduce IFCA-MIR, 
an improved version of IFCA 
that integrates MIA risk assessment into the clustering process. 
Allowing clients to select clusters based on both model performance and MIA vulnerability, IFCA-MIR achieves an improved performance with respect to accuracy, fairness, and privacy. 
We demonstrate that IFCA-MIR significantly reduces MIA risk while maintaining comparable model accuracy and fairness as the original IFCA.
\end{abstract}


\maketitle

\section{Introduction}

Recent advancements in machine learning technologies have positioned artificial intelligence (AI) as a cornerstone of innovation across various sectors, driving progress in healthcare, finance, education, and beyond. However, concerns about social risks, particularly privacy violations, have become increasingly prominent. These concerns underscore the need for privacy-preserving AI solutions capable of safeguarding sensitive data of the clients while maintaining model performance.

Federated Learning (FL)~\cite{mcmahan2017communication} has become a key approach for enabling collaborative model training across multiple clients without the need to share raw data. In FL, participants locally train models on their devices using their own data, transmitting only model updates to a central server. This decentralized approach ensures that raw data remains on clients' devices, effectively minimizing the risk of privacy breaches during the training process.

Despite these advantages as far as privacy is concerned, FL faces significant challenges when applied to real-world scenarios 
due to the non-IID nature of clients' data. In real-world FL applications, clients' data distributions are typically non-IID, reflecting clients' diverse behaviors, preferences, and environments. This heterogeneity of data distribution present in the population impedes model convergence and deteriorates model accuracy. For instance, in healthcare, patient data varies widely across hospitals due to demographic differences, disease prevalence, and medical histories. In this context, a one-size-fits-all approach to obtain a model is inadequate, as it fails to capture the nuanced patterns within each group. Therefore, tailoring models to accommodate these variations is crucial for enhancing predictive accuracy and ensuring fairness across different client groups.

To address the challenge imposed by the non-IID-ness of the training data, Personalized Federated Learning (PFL) approaches have been proposed. These, typically, cluster participants based on similar data distributions and train models tailored to each group. 
A cutting-edge method is the Iterative Federated Clustering Algorithm (IFCA)~\cite{ghosh}, which employs a clustering-based approach for PFL. In IFCA, the server generates multiple models and distributes them to participating clients. Each client then selects the best model that minimizes local loss, performs local training using the selected model, and sends the locally optimized models back to the server. The server aggregates and updates the models, and redistributes the refined models to the clients. This iterative process continues until the model reaches a specified accuracy level or completes a predetermined number of training rounds.

IFCA not only enhances model accuracy compared to traditional FL models but also improves fairness—an increasingly critical concern in AI ethics. Specifically, in scenarios where datasets include both privileged and unprivileged groups, IFCA-based personalized learning demonstrates better performance in accuracy and fairness, as validated by previous studies \cite{galli2023advancing}. Moreover, it has been shown that IFCA can be incorporated with local obfuscation techniques to foster group-level formal privacy guarantees~\cite{galli2023group}.

However, despite these improvements in accuracy, group privacy, and fairness, IFCA has a notable drawback: it becomes increasingly vulnerable to Membership Inference Attacks (MIA) as it partitions the overall dataset into smaller clusters for personalized training. MIA are a significant privacy threat in machine learning, aiming to determine whether a particular data point was used in a model’s training set. In IFCA, this vulnerability is particularly pronounced among unprivileged groups or minorities, where smaller training datasets increase the likelihood of successful MIA.

In this paper, we propose an improved version of IFCA called \sys (Iterative Federated Clustering Algorithm with Membership Inference Robustness) to address these limitations. Unlike the original IFCA, which selects models solely based on empirical loss, \sys integrates MIA risk assessment into the model selection process. By balancing both loss and MIA risk, \sys reduces the exposure to MIA while maintaining comparable accuracy and fairness. This approach allows privacy-sensitive clients to choose safer clusters, effectively mitigating MIA vulnerability without significantly compromising model accuracy. Our method is designed to achieve a better trade-off between accuracy and privacy, addressing the original IFCA’s vulnerability to MIA.

The key contributions of this paper are as follows:
\begin{itemize}
  \item We empirically demonstrate the MIA vulnerability of the original IFCA algorithm, particularly in minority groups with smaller training datasets. To the best of our knowledge, this is the first study to evaluate the MIA vulnerability of clustering-based personalization algorithms.
  \item We propose \sys, an enhanced PFL algorithm that integrates MIA risk into the model selection process, balancing empirical loss with privacy considerations.
  \item We show that \sys preserves the formal convergence guarantees of IFCA and empirically evaluate the performance of \sys with respect to MIA robustness, fairness, and accuracy through extensive experiments on MNIST, FEMNIST, and CIFAR-10 datasets, demonstrating its ability to reduce MIA vulnerability while maintaining comparable model accuracy and fairness.  
\end{itemize}

The rest of this paper is organized as follows: In sections 2 and 3, we introduce related works and preliminaries. Section 4 details the proposed \sys algorithm and analyzes its formal convergence guarantees. Section 5 presents experimental results demonstrating that the proposed algorithm reduces MIA vulnerability without significantly compromising accuracy. Section 6 discusses key insights that require further exploration. Finally, Section 7 concludes the paper and outlines directions for future work.

\section{Related works}
\subsection{Federated learning}
FL is a framework for collaborative model training across multiple clients without requiring the sharing of raw data. This decentralized approach preserves data privacy by ensuring that raw data remains on clients’ devices, thereby reducing the risk of data leakage during training.

However, despite its inherent privacy-preserving design, FL is still vulnerable to various privacy attacks, as adversaries can exploit shared model updates to extract sensitive information. Studies have demonstrated that malicious clients, servers, or external attackers can utilize gradient inversion techniques \cite{geiping2020inverting}, differential data leakage attacks \cite{dlg}, and source inference attacks \cite{hu2021source} to reconstruct private training data or infer membership. These vulnerabilities reveal the limitations of standard FL protocols in safeguarding client privacy, emphasizing the need for enhanced privacy-preserving mechanisms.

In addition to privacy concerns, one of the critical challenges in FL is the presence of non-IID data across clients. Several lines of work~\cite{ghosh, sattler, mansour2020three,tan2022towards} have begun exploring PFL approaches, which aim to tailor models to individual clients’ unique data distributions. However, these methods primarily focus on improving model accuracy and convergence without adequately addressing the accompanying privacy risks. In particular, the non-IID nature of data amplifies the risk of privacy attacks, as attackers can exploit unique data distributions to infer sensitive information. Therefore, there is a growing need for PFL algorithms that not only address non-IID data issues but also incorporate privacy-preserving measures.

\subsection{Membership inference attack}
MIA \cite{shokri2017membership,hu2022membership,choquette2021label} are a class of privacy attacks in which an adversary attempts to determine whether a specific data sample was used in training a machine learning model. These attacks exploit differences in the model’s behavior when processing training and non-training data, often by leveraging confidence scores or loss values. 

Shokri et al. \cite{shokri2017membership} introduced the first large-scale membership inference attack, which utilizes shadow models to approximate the target model's behavior. The attack framework consists of the following steps:
\begin{itemize}
    \item \textbf{Shadow Model Training:} The adversary trains multiple shadow models that mimic the behavior of the target model using auxiliary data.
    \item \textbf{Attack Model Construction:} An attack model is trained using the outputs of shadow models, learning to distinguish between samples that were used for training and those that were not.
    \item \textbf{Inference Phase:} The trained attack model is applied to the target model’s outputs, classifying input samples as either "member" (included in training) or "non-member" (not included in training).
\end{itemize}

Numerous studies have shown that overfitting, small training set sizes increase susceptibility to MIA \cite{shokri2017membership, tobaben2024impact}. Specifically, MIA has been shown to be effective even in federated settings, where raw data is not shared but model updates are still vulnerable to inference attacks \cite{hu2022membership}. Moreover, the clustering strategy in IFCA exacerbates MIA risks, as splitting data into smaller clusters makes minority groups more vulnerable to membership inference.

To counter MIA, various defense mechanisms have been proposed, including differential privacy (DP) \cite{abadi2016deep}, adversarial regularization \cite{nasr2018machine}, and knowledge distillation \cite{shejwalkar2021membership}. However, these methods often come at the cost of reduced model accuracy and are not specifically designed to address the unique challenges of PFL. Consequently, there is a pressing need for privacy-preserving PFL frameworks that balance accuracy and privacy.

\section{Preliminaries}
This section provides the foundational background necessary to understand the proposed \sys algorithm, including an overview of PFL using IFCA, fairness metrics in machine learning, and the MIA accuracy metric used to evaluate privacy risks.

\subsection{Personalized FL with IFCA}

In IFCA, the learning problem is framed as a stochastic optimization problem. The objective is to find a set of optimal model parameters $\theta_j^* \in \mathbb{R}^d$ for each cluster $j \in [s]$ such that
\begin{equation} \label{erm:1} F(\theta_j) = \mathbb{E}_{z\sim\mathcal{D}_j} \left[f(\theta_j;z)\right], \end{equation} 
where  $f(\theta_j;z)$ is  is the local loss function evaluated on data point $z$ under model parameters $\theta_j$ and ${\Pi_j}$ denotes the underlying data distributions for cluster $j$ which are not directly accessible.  Instead, they are accessed through a client datasets $Z_c=\left\{z | z \sim \Pi_j, z \in \mathbb{D} \right\}$ where $c$ denotes a client and $\mathbb{D}$ is the domain of data points. The goal is to estimate the membership of each client $c$ to one of the clusters and minimize the empirical loss:

\begin{equation} \tilde{F}(\theta_j) = \frac{1}{|S_j|}\sum_{c \in S_j} \tilde{F_c}(\theta_j; Z_c), \end{equation}

\begin{equation} \tilde{F_c}(\theta_j; Z_c) = \frac{1}{|Z_c|}\sum_{z_i \in Z_c} f(\theta_j; z_i), \end{equation}

where $S_j$  denotes the set of clients assigned to cluster 
$j$. The optimization objective is to find the optimal model parameters for each cluster:
\begin{equation}
\tilde{\theta}{j}^* = \argmin_{\theta_j}\tilde{F}(\theta_j).
\end{equation}

\subsection{Fairness}
Fairness has emerged as a critical concern in machine learning, especially in federated learning scenarios where data is distributed across diverse client groups \cite{pessach2022review, ezzeldin2021fairfed,wick2019unlocking,menon2018cost}. In this study, we evaluate fairness using three commonly accepted metrics: Demographic Parity \cite{dwork2012fairness}, Equal Opportunity \cite{hardt2016equality}, and Equalized Odds \cite{hardt2016equality}. We use the notation $\hat{Y}=1$, $\hat{Y}=0$ to indicate positive and negative predictions, respectively, and $S=1$, $S=0$ to denote the privileged and unprivileged groups.

\begin{definition}\label{def:demoparity}
\emph{Demographic Parity} is requires that the model’s predictions be independent of sensitive attributes. Formally, it is defined as:
\begin{equation}
\mathbb{P}\left[\hat{Y}=1| S=1\right]=\mathbb{P}\left[\hat{Y}=1| S=0\right]
\end{equation}
\end{definition}

\begin{definition}\label{def:eqaulopp} \emph{Equal Opportunity}  ensures that true positive rates are equal across groups, promoting fairness without sacrificing accuracy:
\begin{equation}
    \mathbb{P}\left[\hat{Y}=1|Y=1, S=1\right]=\mathbb{P}\left[\hat{Y}=1|Y=1, S=0\right]
\end{equation}
\end{definition}

\begin{definition}\label{def:eqaulodd} \emph{Equalized Odds} 
 require equal true positive rates and false positive rates across groups:
\begin{equation}
\mathbb{P}\left[\hat{Y}=1|Y=y, S=1\right]=\mathbb{P}\left[\hat{Y}=1|Y=y, S=0\right] \end{equation}
This ensures balanced accuracy across both positive and negative outcomes.
\end{definition}

In practice, achieving perfect equality in these metrics is challenging. Therefore, the goal is to minimize the absolute difference between the privileged and unprivileged groups, ensuring equitable model performance.

\subsection{Merbership inference attack accuracy}
We measure MIA accuracy using the following metric:

\begin{equation}
\text{MIA Accuracy} = \frac{\text{TPR} + \text{TNR}}{2}
\end{equation}

where the True Positive Rate (TPR) and True Negative Rate (TNR) are defined as follows:

\begin{equation}
\text{TPR} = \frac{|\{ x \in D_{\text{train}} \mid \hat{y}(x) = 1 \}|}{|D_{\text{train}}|}
\end{equation}

\begin{equation}
\text{TNR} = \frac{|\{ x \in D_{\text{non-train}} \mid \hat{y}(x) = 0 \}|}{|D_{\text{non-train}}|}
\end{equation}

Here, \( D_{\text{train}} \) represents the set of training samples, and \( D_{\text{non-train}} \) represents the set of non-training samples. The function \( \hat{y}(x) \) denotes the MIA classifier's prediction, where \( \hat{y}(x) = 1 \) indicates that the sample is inferred as part of the training set, and \( \hat{y}(x) = 0 \) indicates it is inferred as non-training data.

\section{Iterative Federated Clustering Algorithm with Membership Inference Robustness}
\subsection{Motivation}
As mentioned from the above, while IFCA offers advantages in terms of both accuracy and fairness, it becomes more vulnerable to MIA. This vulnerability arises from IFCA's clustering strategy, where the entire training dataset is divided into smaller groups, and models are trained separately for each group. The accuracy of MIA is inversely proportional to the size of the training dataset, with minority groups being particularly more weak to such attacks.

\begin{figure}[htbp]
\centerline{\includegraphics[width=0.5\textwidth]{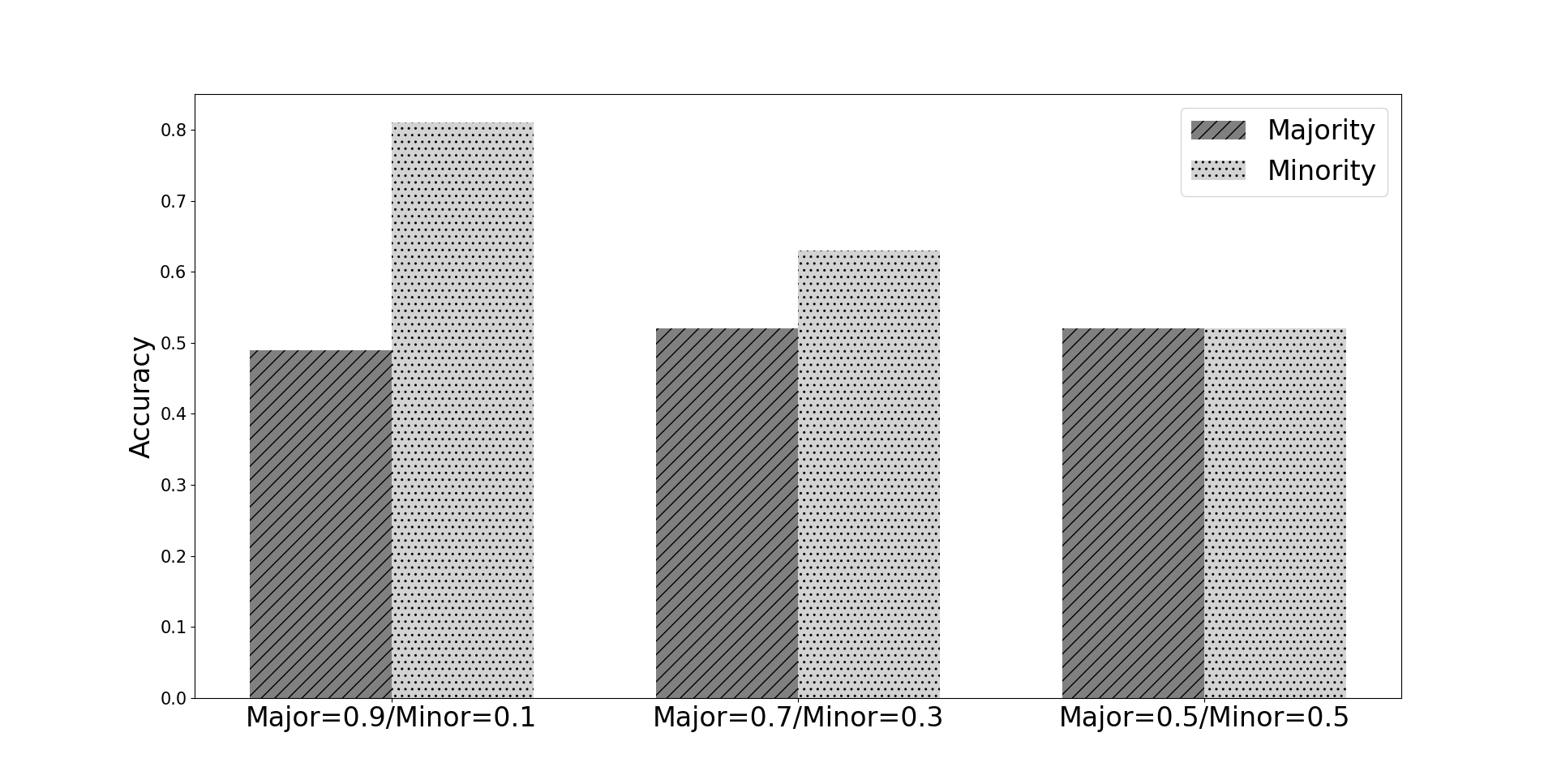}}
\caption{MIA accuracy according to training dataset size}
\label{fig:mia_accuracy}
\end{figure}

Figure \ref{fig:mia_accuracy} shows the results of performing MIA using Shokri et al.'s method \cite{shokri2017membership} after applying the IFCA technique to the MNIST dataset. Following the experimental setup in \cite{galli2023advancing, galli2023group}, we configured the MNIST dataset into majority and minority groups with different distributions. When the proportion of minority participants was set to 10\% out of the 200 total participants, the MIA accuracy for the minority group reached 81\%, whereas the MIA accuracy for the majority group was 51.5\%. When the minority group constituted 30\% of the participants, the MIA accuracy dropped to 63\%, and when both groups had an equal 50\% split, the MIA accuracy was 52\%. This demonstrates that MIA accuracy does not significantly increase beyond a certain dataset size, but for datasets below that threshold, the smaller the group, the more vulnerable it becomes to MIA. In other words, when applying IFCA for personalized FL, clients belonging to the minority group are exposed to a higher risk of MIA.

To address this disadvantage, we propose an improved version of IFCA, called \sys, which incorporates MIA accuracy into the cost function alongside empirical loss. This new algorithm allows personalized models to be trained for each group, minimizing the risk of MIA while maintaining model performance.

\subsection{Notations and problem formulation}\label{subsec:notations} 
\subsubsection*{Notations}
Let $\mathcal{C}=\{C_1,\ldots,C_n\}$ be a set of $n$ clients that wish to collaboratively train a model in a federated environment in the presence of an aggregating entity (e.g., a server). Let the clients in $\mathcal{C}$ be holding data that are partitioned into $s$ distributions $\Pi_1,\ldots\Pi_s$ for some $s\leq n$. For every $i\in [n]$ and $j\in[s]$, let $Z_i$ be the dataset held by $C_i$ where the datapoints are sampled from the space $\mathcal{Z}$ and let $S^*_{[j]}$ be the set of clients whose data are sampled from $\Pi_j$, $i.e., S^*_{[j]}=\{i:\, z\sim \Pi_j\,\forall\,z\in Z_i\}$. Finally, in any round $t$, let $\theta_{c[j]}^{(t)}$ denotes the local model of $c\in\mathcal{C}$ optimized for cluster $j$ and let $S^{(t)}_{[j]}$ be the set of all clients in round $t$ who reported their optimal cluster ID as $j$. Hence, let $\theta^{(t+1)}_{[j]}=\frac{1}{\left\lvert S^{(t)_{[j]}}\right\rvert}\sum_{c\in S^{(t)}_{[j]}}\theta_{c[j]}^{(t)}$ denote the aggregated optimized model for cluster $j$ in round $t$.

\subsubsection*{Problem formulation} 
Let $f:\mathbb{R}^d\times \mathcal{Z}\mapsto \mathbb{R}_{\geq 0}$, where $d$ is the dimension of the model space, be the loss function that the clients seek to minimize for each cluster by finding an optimal model. 

In practice, each client holding data sampled from one of the $s$ distributions uses a small sample of their datapoints to evaluate and minimize the average loss they incur on the model in each round to update their local model. Thus, for any finite \emph{batch} $\xi\subset \mathcal{Z}$, the average loss on that batch $\xi$ for any model 
$\theta\in\mathbb{R}^d$ is given by $f(\theta,\xi)=\frac{1}{|\xi|}\sum_{z\in \xi}f(\theta,z)$. Moreover, let $\ell(\theta,Z)$ denote the accuracy of MIA on $\theta$ for any training dataset $Z$. Therefore, the primary training objective of \sys can be formally expressed as:
\begin{align}
    &\forall\,j\in [s]: \nonumber\\
    &\theta^*_{[j]}=\argmin_{\theta\in\mathbb{R}^d}{\sum_{i\in S^*_{[j]}}{\frac{1}{|S^*_{[j]}|}}(\alpha_if(\theta,Z_i)+\beta_i \ell (\theta,Z_i))},\label{eq:prob_formulation}
\end{align}
where, for every $i\in[n]$, $\alpha_i\in [0, 1]$ and $\beta_i\in [0, 1]$ with $\alpha_i+\beta_i=1$ are hyperparameters chosen by $C_i$ based on its preference on accuracy or privacy. 

\begin{remark}
   A high value of $\alpha_i$ and a low value of $\beta_i$ would exemplify the case where $C_i$ has a greater priority for its model accuracy and less for being protected from a potential MIA. An extreme case of $\alpha_i=1$ and $\beta_i=0$ for every $i\in [n]$ reduces down to IFCA.  
\end{remark}

\subsection{Proposed Method: \sys}
The core idea of \sys is that the cluster selection considers not only the empirical loss but also the MIA accuracy. For instance, supposed a client $C_i$ has a local dataset $z_i$ and must choose between models $\theta_j$ and $\theta_k$. Even if $\theta_j$ yields a lower empirical loss on $z_i$, it may also exhibit a higher MIA accuracy, indicating greater vulnerability to membership inference attacks. In such a case, a privacy-sensitive client might prefer $\theta_k$, which has a higher empirical loss but offers better privacy protection.

Although \sys cannot directly reduce the risk of MIA itself, it enables clients to balance privacy protection and model performance by considering both privacy preferences and empirical loss during cluster selection. To achieve this, we propose the following enhancements in the \sys algorithm:
\begin{itemize}
\item
Red team role for MIA evaluation: The server acts as a red team, simulating external attackers by performing MIA on each $\theta$. In this process, the server calculates the MIA accuracy to assess the privacy vulnerability of each $\theta$, indicating how susceptible each $\theta$ is to MIA.
\item
Providing privacy evaluation information: When the server distributes the $\theta$ to clients, it provides not only the candidate $\theta$ but also the MIA accuracy information for each $\theta$. Clients can use this information to assess the privacy risks associated with each $\theta$.
\item
Client's optimal cluster selection: Clients evaluate the empirical loss of each $\theta$ received from the server, as done in the original IFCA. However, in \sys, clients also consider the MIA accuracy of each $\theta$ based on their privacy sensitivity. This approach allows each client to select the $\theta$ that strikes the best balance between privacy and performance for their specific needs.
\end{itemize}

To implement these improvements, we propose the following \sys algorithm.

\begin{algorithm2e}[t]
    \DontPrintSemicolon
    \caption{\sys }
    \label{alg:IFCAMIR}
    
    \KwIn{Clients $\mathcal{C} = \{C_1, \dots, C_n\}$, Initial models $\{\theta^{(0)}_{[1]}, \dots, \theta^{(0)}_{[s]}\}$, Learning rate $\eta$, Number of rounds $T$}
    \KwOut{Optimized models $\{\theta^{(T)}_{[1]}, \dots, \theta^{(T)}_{[s]}\}$}
    
    \tcp{Server-side}
    Randomly initialize $\theta^{(0)}_{[1]}, \dots, \theta^{(0)}_{[s]}$\;
    
    \For{$t = 0, \dots, T-1$}{
        \tcp{Train MIA models and compute MIA accuracy}
        \For{$j = 1, \dots, s$}{
            Compute MIA risk $\ell(\theta^{(t)}_{[j]})$ using \textbf{TrainMIA}($\theta^{(t)}_{[j]}$)\;
        }
        
        Send $\{(\theta^{(t)}_{[1]}, \ell(\theta^{(t)}_{[1]})), \dots, (\theta^{(t)}_{[s]}, \ell(\theta^{(t)}_{[s]}))\}$ to selected clients\;
        
        \tcp{Client-side}
        \ForEach{Client $C_i \in \mathcal{C}$ in parallel}{
            Sample mini-batch $\xi_i$ from local dataset $Z_i$\;
            
            \tcp{Select best model based on empirical loss and MIA risk}
            $\hat{j}_i = \argmin_{j \in [s]} \alpha_i f(\theta^{(t)}_{[j]}, \xi_i) + \beta_i \ell(\theta^{(t)}_{[j]})$\;
                $\tilde{\theta}^{(t)}_{[ \hat{j}_i]} \leftarrow \tilde{\theta}^{(t)}_{[ \hat{j}_i]} - \eta \nabla f(\tilde{\theta}^{(t)}_{[ \hat{j}_i]}, \xi_i)$\;
            Send $(\tilde{\theta}^{(t)}_{[ \hat{j}_i]}, \hat{j}_i)$ to server\;
        }
        \tcp{Server-side Aggregation}
        \For{$j = 1, \dots, s$}{
            $\theta^{(t+1)}_{[j]}=\frac{1}{\left\lvert S^{(t)_{[j]}}\right\rvert}\sum_{c\in S^{(t)}_{[j]}}\theta_{c[j]}^{(t)}$\;
        }
    }
    \KwRet Optimized models $\{\theta^{(T)}_{[1]}, \dots, \theta^{(T)}_{[s]}\}$
\end{algorithm2e}

\begin{algorithm2e}[t]
    \DontPrintSemicolon
    \caption{TrainMIA}
    \label{alg:TrainMIA}
    
    \KwIn{model $\theta_{[j]}^{(t)}$, Shadow dataset $D_{\text{shadow}}$}
    \KwOut{MIA risk score $\ell(\theta_{[j]}^{(t)})$}
    
    Train multiple shadow models $\{f_{\theta_{[j]}^{(t)}}^{\text{shadow}}\}$ on $D_{\text{shadow}}$\;
    
    \tcp{Construct attack dataset}
    \ForEach{sample $z$ in $D_{\text{shadow}}$}{
        Compute confidence scores $S_z = f_{\theta_{[j]}^{(t)}}^{\text{shadow}}(z)$\;
        Assign label $y = 1$ if $z \in D_{\text{train}}^{\text{shadow}}$, else $y = 0$\;
        Store $(S_z, y)$ in attack dataset\;
    }
    Train attack model $A$ on the attack dataset $(S_z, y)$\;

    Compute MIA accuracy $\ell(\theta_{[j]}^{(t)}) = \frac{\text{TPR} + \text{TNR}}{2}$\;
    
    \KwRet $\ell(\theta_{[j]}^{(t)})$
\end{algorithm2e}

There are two main differences from the original IFCA in this algorithm. First, in \sys, the server performs MIA on each model $\theta$ (Algorithm \ref{alg:TrainMIA}) and provides the MIA accuracy information to clients, enabling them to make more informed decisions (line 7 in Algorithm \ref{alg:IFCAMIR}). Second, clients select the optimal $\theta^*$ based on both the empirical loss and the MIA accuracy of the provided $\theta$ (line 12 in Algorithm  \ref{alg:IFCAMIR}).

\subsection{Convergence analysis}\label{subsec:convergence}
In this section, we present the theoretical analysis of convergence guarantees of \sys. For this, in addition to the notations introduced in \Cref{subsec:notations}, we define the following additional terms. For any model $\theta\in\mathbb{R}^d$ and $j\in[s]$, let $F_{[j]}(\theta)$ denote the \emph{population loss} of cluster $j$ for model $\theta$ and is given by the expected loss of $\theta$ over the data points following the distribution $\Pi_j$, \ie $F_{[j]}(\theta)=\mathbb{E}_{z\sim \Pi_j}\left[f(\theta,z)\right]$. For any client $C_i\in\mathcal{C}$ and model $\theta\in\mathbb{R}^d$, we call $\hat{f}_i(\theta,z)=\alpha_i f(\theta,z) +\beta_i (\theta) \ell(\theta)$ the \emph{privacy-aware loss} of client $C_i$ for model $\theta$ computed on data point $z$. Correspondingly, for all $j\in[s]$ let $\hat{F}_{[j]}$ denote the \emph{population privacy-aware loss} of cluster $j$ given by the average privacy-aware loss of the clients in $S^*_{[j]}$, \ie,
$$\hat{F}_{[j]}(\theta)=\frac{1}{\left\lvert S^*_{[j]}\right\rvert}\sum_{i\colon C_i\in S^*_{[j]}}\mathbb{E}_{z\sim \Pi_j}\left[\hat{f}_i(\theta,z)\right].$$ 
Let $\Delta$ denote the \emph{minimum difference} between the optimal models of any two clusters, \ie, $\Delta=\min_{j\neq j'}\norm{\theta^{*}_{[j]}-\theta^{*}_{[j']}}$. Finally, for every $j=1,\ldots, s$, let $p_j=\left\lvert S^{*}_{[j]}\right\rvert/n$ denote the fraction of clients belonging to $S^*_{[j]}$ and, hence, $p=\min_{j=1,\ldots,s}p_j$.

In order to proceed with the convergence analysis of \sys, we let each client in any given round use a batch of size $B<\infty$ of their local data points to run their local training and we adhere to the same set of assumptions on the loss functions and the initialization conditions that the formal convergence guarantees of IFCA rely on~\cite{ghosh}.

\begin{assumption}[Strong convexity and smoothness of the population privacy-aware loss function]\label{assump:strong_convex_L_smooth}
    For each $j\in[s]$, the corresponding population loss function $\hat{F}_{[j]}$ is $\lambda$\emph{-strongly convex} and $L$\emph{-smooth}.
\end{assumption}

\begin{assumption}[Bounded variance of privacy-aware loss function]\label{assump:bounded_var_loss}
    For any $\theta\in\mathbb{R}^d$, $j\in[s]$, and $i\in[n]$, the variance of $\hat{f}_i(\theta,z)$ is bounded by $\eta^2$ where $z\sim\Pi_j$, \ie, $\mathbb{E}_{z\sim \Pi_j}\left[(\hat{f}_i(\theta,z)-\hat{F}_{[j]}(\theta,z))^2\right]\leq \eta^2$.
\end{assumption}

\begin{assumption}[Bounded variance on gradients]\label{assump:bounded_var_grad}
    For any $\theta\in\mathbb{R}^d$, $j\in[s]$, and $i\in[n]$, the variance of $\nabla \hat{f}_i(\theta,z)$ is bounded by $\sigma^2$ where $z\sim\Pi_j$, \ie, $\mathbb{E}_{z\sim \Pi_j}\left[\norm{\nabla \hat{f}_i(\theta,z)-\nabla \hat{F}_{[j]}(\theta,z)}^2_2\right]\leq \sigma^2$.
\end{assumption}

\begin{assumption}[Initialization condition]\label{assump:init}
    Without the loss of generality, let $\max_{j\in [s]}\norm{\theta^*_{[j]}}\leq 1$. Then for every cluster $j\in[k]$, we assume \sys to satisfy the following initialization conditions:
    \begin{align}
         &\norm{{\theta}^{(0)}_{[j]}-\theta^{*}_{[j]}}\leq (\frac{1}{2}-\alpha)\sqrt{\frac{\lambda}{L}}\Delta ,\nonumber\\
         & \text{such that $0\leq\alpha \leq\frac{1}{2}$},\,B\geq \frac{s \eta^2}{\alpha^2\lambda^2\Delta^4},\, p\geq \frac{\log(nB)}{n},
         \text{ and}\nonumber\\
         &\Delta \geq \tilde{\mathcal{O}}\left(\max\{\alpha^{-2/5}B^{-1/5},\,\alpha^{-1/3}n^{-1/6}B^{-1/3}\}\right).\nonumber
     \end{align}
     where $\tilde{\mathcal{O}}$ any logarithmic factors and terms that do not depend on $n$ and $B$.
\end{assumption}

 \begin{theorem}\label{th:local_convergence}
     If Assumptions~\ref{assump:strong_convex_L_smooth},\ref{assump:bounded_var_loss},\ref{assump:bounded_var_grad}, and \ref{assump:init} hold, choosing learning rate $\eta=1/L$, each cluster $j\in [s]$, and any $\delta\in(0,1)$, in every round $t>0$, we have with probability at least $(1-\delta)$:
     \begin{align}
         \norm{{\theta}^{(t+1)}_{[j]}-\theta^{*}_{[j]}}\leq \left(1-\frac{p\lambda}{8L}\right) \norm{{\theta}^{(t)}_{[j]}-\theta^{*}_{[j]}}+\epsilon_0\nonumber
     \end{align}
     where $\epsilon_0 \leq \frac{\sigma}{\delta L \sqrt{pnB}}+\frac{\eta^2}{\delta \alpha^2 \lambda^2 \Delta^4 B}+\frac{\eta\sigma s^{3/2}}{\delta^{3/2}\alpha \lambda L \Delta^2 \sqrt{n}B}$.
 \end{theorem}

 \begin{corollary}\label{th:final_convergence}
    If Assumptions~\ref{assump:strong_convex_L_smooth},\ref{assump:bounded_var_loss},\ref{assump:bounded_var_grad}, and \ref{assump:init} hold, choosing learning rate $\eta=1/L$, for each cluster $j\in [s]$ in every round $t>0$, and any $\delta\in (0,1)$ and any $\epsilon>0$, setting $\hat T=\frac{8L}{p\lambda}\log(\frac{2\Delta}{\epsilon})$, in any round $t\geq \hat T$ of \sys, we have with probability at least $(1-\delta)$:
     \begin{align}
         &\norm{\theta^{(t)}_{[j]}-\theta^{*}_{[j]}}\leq\epsilon\nonumber
     \end{align}
     where $\epsilon \leq \frac{\sigma s L \log(nB)}{p^{5/2}\lambda^2 \delta \sqrt{nB}}+\frac{\eta^2 L^2 s \log(nB)}{p^2 \lambda^4 \delta \Delta^4 B}+\tilde{\mathcal{O}}\left(\frac{1}{B\sqrt{n}}\right)$.
 \end{corollary}

The proofs of \Cref{th:local_convergence} and \Cref{th:final_convergence} follow directly from the reasoning presented in the proofs of Theorem 2 and Corollary 2 in \cite{ghosh}.

\section{Experiments}
In this section, we evaluate the proposed \sys method and analyze its effectiveness in mitigating MIA risks while maintaining model accuracy and fairness. The experiments were conducted on the MNIST, FEMNIST, and CIFAR-10 datasets, comparing the performance of \sys against the original IFCA algorithm.

The objectives of our evaluation are twofold:
\begin{itemize}
    \item To assess whether \sys can reduce MIA vulnerability without compromising model accuracy.
    \item To examine whether the fairness of the federated learning model is preserved when incorporating MIA robustness into model selection.
\end{itemize}

By addressing both the privacy and model accuracy aspects in our evaluation, this experiment provides a comprehensive analysis of the \sys algorithm's effectiveness in real-world PFL scenarios.

\subsection{Datasets}\label{AA}
\subsubsection{MNIST}

The MNIST dataset is a widely-used benchmark for handwritten digit recognition tasks. It consists of grayscale images representing digits 0 through 9, with each image standardized to a size of 28×28 pixels. The dataset includes 60,000 training images and 10,000 test images.

For our experiments, we distributed 50,000 training images evenly among 200 clients for PFL training, while the remaining 10,000 images were used to train a shadow model for MIA analysis on the server.

We preprocessed MNIST based on prior methodologies from \cite{ghosh} and \cite{galli2023advancing}:
\begin{itemize}
    \item \cite{ghosh}: Rotated MNIST images at different angles to validate personalization.
    \item \cite{galli2023advancing}: Segmenting the dataset into privileged and unprivileged groups to estimate the fairness.
\end{itemize}
To simulate real-world non-IID settings, we introduced image rotation variations per client uniformly rather than applying fixed rotation per group:
\begin{itemize}
\item Minority group rotations: (0–20°), (0–25°), and (0–30°).
\item Majority group rotations: (20–40°), (25–50°), and (30–60°).
\end{itemize}

By adjusting the distance between the angle boundaries of the two groups, we aim to more accurately simulate real-world data heterogeneity, where distributions often overlap rather than having clearly defined boundaries.

\subsubsection{FEMNIST}
The FEMNIST (Federated Extended MNIST) dataset is specifically designed to evaluate performance in federated learning environments and exhibits a non-IID data distribution \cite{caldas2018leaf}. Derived from the EMNIST dataset, FEMNIST consists of handwritten images representing 62 classes, including digits 0-9 and both uppercase and lowercase alphabetic characters (A-Z, a-z). The dataset comprises approximately 700,000 training samples and 100,000 test samples, collected from around 3,500 clients. 

To control the inherent non-IID characteristics of the FEMNIST dataset, we first aggregated the training data, which was originally divided by individual clients, into a unified dataset. We then evenly distributed 600,000 training samples among 200 clients and used the remaining 100,000 samples to train a shadow model for MIA on the server.

We preprocessed the FEMNIST dataset using the same methodology as that used for the MNIST dataset.
\begin{itemize}
\item Minority group rotations: (0–20°), (0–25°), and (0–30°).
\item Majority group rotations: (25–45°), (30–55°), and (35–65°).
\end{itemize}

\subsubsection{CIFAR-10}
The CIFAR-10 dataset consists of 60,000 color images categorized into 10 classes, with each image sized at 32x32 pixels. It contains 50,000 training images and 10,000 test images, uniformly distributed across all classes. For our study, 50,000 training images were evenly allocated to 100 clients for PFL training and 10,000 images were used to train the shadow model for MIA.

Since CIFAR-10 images are more complex than those in MNIST and FEMNIST, controlling the non-IID characteristics through image rotation was not feasible. Instead, we varied the brightness levels to simulate non-IID data distributions. 
\begin{itemize}
\item Minority group rotations: (0.5-0.8)
\item Majority group rotations: (0.8-1.1), (0.85-1.15), and (0.9-1.2).
\end{itemize}

This approach introduces non-IID conditions while preserving image integrity.

\subsection{Experimental Setup }
The goal of our experiment is to verify whether the \sys can minimize the number of clients exposed to MIA risks without compromising model accuracy, while also maintaining fairness. To achieve this, we designed experiments focusing on three key aspects: model accuracy, MIA vulnerability, and fairness. Each experiment was repeated five times to ensure reliability. Additionally, the server performs MIA evaluations every five iterations with three shadow models rather than at every iteration. The frequency of MIA evaluations and number of shadow models can be adjusted as needed based on specific requirements. 

\subsubsection{Model Accuracy}
To evaluate model accuracy, we used the MNIST, FEMNIST, and CIFAR-10 datasets, comparing the performance of \sys against the original IFCA algorithm.

For MNIST and FEMNIST, the total number of clients was set to 200, with each client receiving an equal number of data points. For CIFAR-10, the number of clients was set to 100.

The objective was to compare model accuracy between the majority and minority groups. We assumed the presence of two models for PFL, aligning with the two distinct distributions in each dataset. we compared the accuracy of the original IFCA algorithm with that of \sys, examining whether \sys could maintain model accuracy without degradation.

\subsubsection{MIA vulnerability}
To evaluate MIA vulnerability, we used the same datasets and experimental settings as those employed in the model accuracy evaluation. In this evaluation, each client was assigned an MIA vulnerability threshold, and we counted the number of clients whose personalized model's MIA accuracy exceeded these thresholds.

The MIA threshold defines the maximum allowable MIA accuracy, reflecting each client's privacy preference. We uniformly assigned the MIA thresholds between 50\% and 80\% across all clients. A lower threshold indicates a higher sensitivity to MIA, while a higher threshold suggests a greater tolerance for MIA risks.

For example, if a client sets an MIA threshold of 60\% but is assigned a personalized model with an MIA accuracy of 65\%, their threshold would be exceeded. Conversely, a client with a threshold of 70\% in the same group would remain within their acceptable limit. This approach enabled us to effectively evaluate the extent to which each client's privacy preferences were upheld.

\subsubsection{Fairness}
To evaluate fairness, we utilized three metrics: demographic parity, equal opportunity, and equalized odds. We compared the results between the majority and minority groups to assess whether fairness was preserved in the proposed method. This analysis allowed us to determine whether the \sys algorithm maintains fairness while simultaneously mitigating MIA vulnerability compared to the original IFCA algorithm.

\subsection{Model Accuracy}

We evaluated the model accuracy of both the original IFCA and the proposed \sys by varying image rotation degrees (for MNIST and FEMNIST) and brightness levels (for CIFAR-10), as well as by adjusting the proportion of the minority group relative to the majority group. Specifically, we compared the accuracy of personalized models for both the majority and minority groups under the original IFCA and \sys methods.

\begin{figure}[htbp]
  \centering
  \begin{subfigure}[b]{0.5\textwidth}
    \centering
    \includegraphics[width=\textwidth]{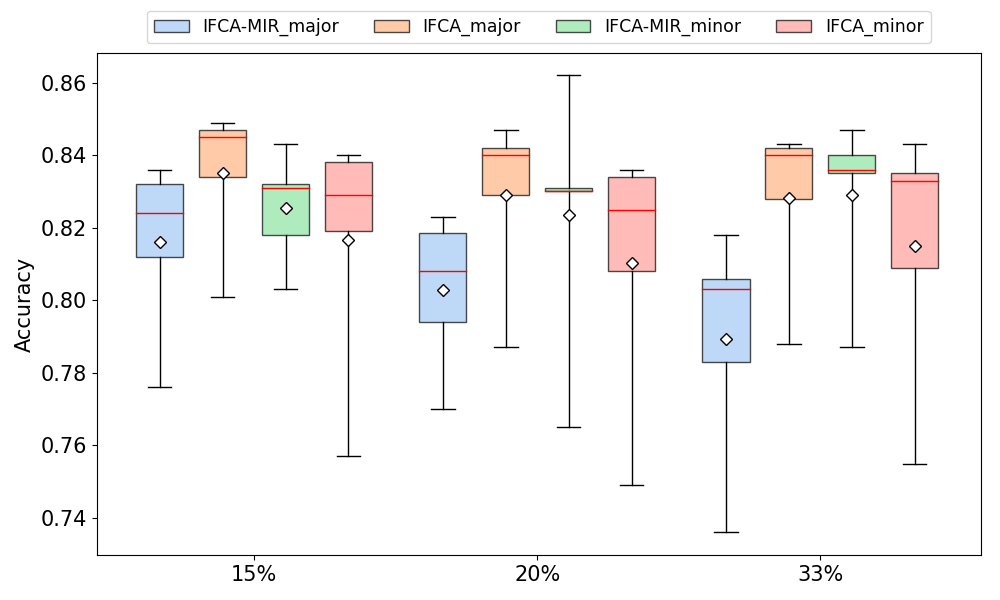}
    \caption{MNIST}
    \label{fig:mnist_num_acc}
  \end{subfigure}
  \hspace{0.001\textwidth} 
  \begin{subfigure}[b]{0.5\textwidth}
    \centering
    \includegraphics[width=\textwidth]{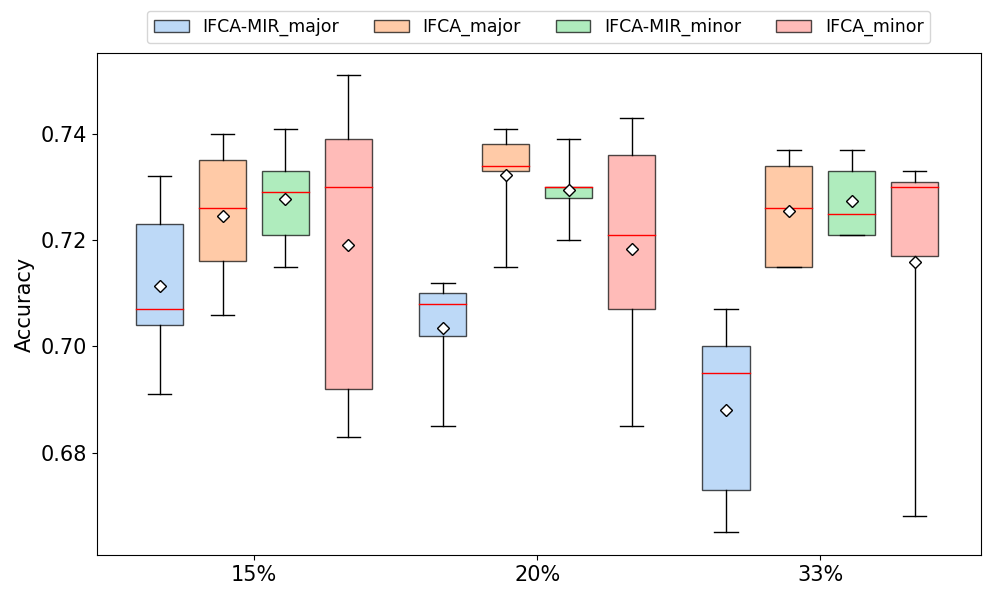}
    \caption{FEMNIST}
    \label{fig:femnist_num_acc}
  \end{subfigure}
  \hspace{0.001\textwidth} 
  \begin{subfigure}[b]{0.5\textwidth}
    \centering
    \includegraphics[width=\textwidth]{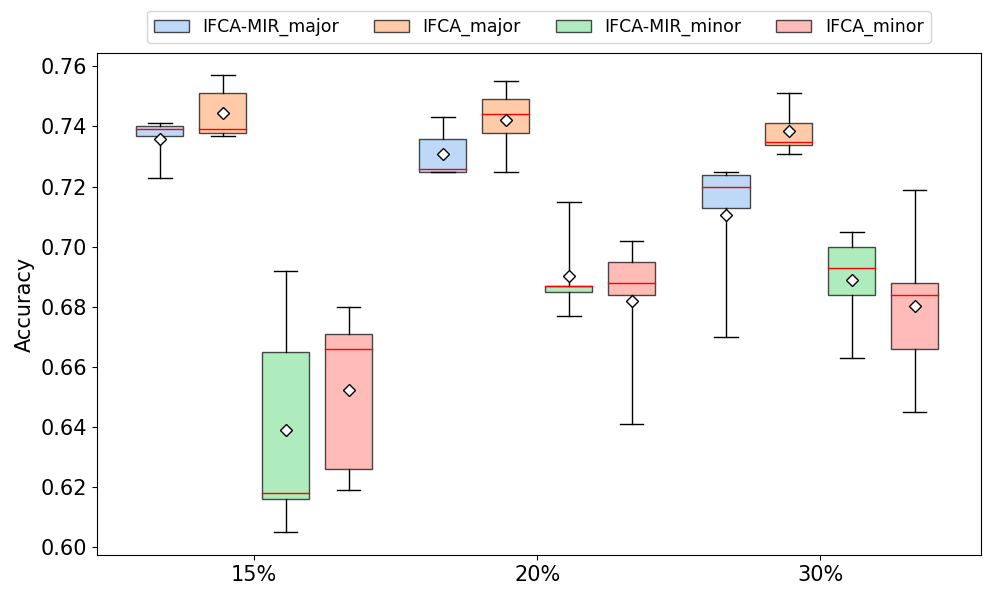}
    \caption{CIFAR-10}
    \label{fig:cifar10_num_acc}
  \end{subfigure}
  \caption{Model accuracy results for each dataset with varying minority dataset sizes. The white diamond represents the mean, while the red line within the box plot indicates the median}
  \label{fig:num_acc}
\end{figure}

Figure \ref{fig:num_acc} illustrates the results when the proportion of the minority group was varied. For MNIST, the majority group’s image rotation angles were fixed within the range of (25-50) degrees, while the minority group’s angles were set within (0-25) degrees. For FEMNIST, the majority group’s rotation angles were fixed within (30-55) degrees, while the minority group’s angles were within (0-25) degrees. In CIFAR-10, the majority group’s brightness was set within the range of (0.85-1.15), whereas the minority group’s brightness was fixed within (0.5-0.8).

The results indicate that for the majority group, \sys yielded slightly lower accuracy compared to the original IFCA. However, this difference was not substantial enough to indicate a significant performance drop. In contrast, the minority group’s accuracy was either comparable to or improved over the original IFCA. This improvement can be attributed to \sys’s incorporation of both MIA vulnerability and loss, which led some clients originally classified as minority to migrate to the majority group to reduce their MIA risks.

As a result, the remaining minority group consisted of clients who were willing to accept a higher MIA risk in exchange for better model accuracy. Consequently, their accuracy either remained equivalent to or exceeded that of the original IFCA. This trend was consistently observed across all datasets, demonstrating that \sys maintains stable performance across different data distributions. However, as shown in Figure \ref{fig:cifar10_num_acc}, when the minority group constituted only 15\% of the total in the CIFAR-10 dataset, the accuracy of the minority model under \sys was lower than that of the original IFCA. This result can be explained by the complexity of CIFAR-10, which demands a larger dataset to achieve high accuracy. In this scenario, despite some clients accepting higher MIA risks, the minority model in \sys exhibited slightly lower accuracy than the original IFCA minority model due to the limited amount of available training data.

\begin{figure}[htbp]
  \centering
  \begin{subfigure}[b]{0.5\textwidth}
    \centering
    \includegraphics[width=\textwidth]{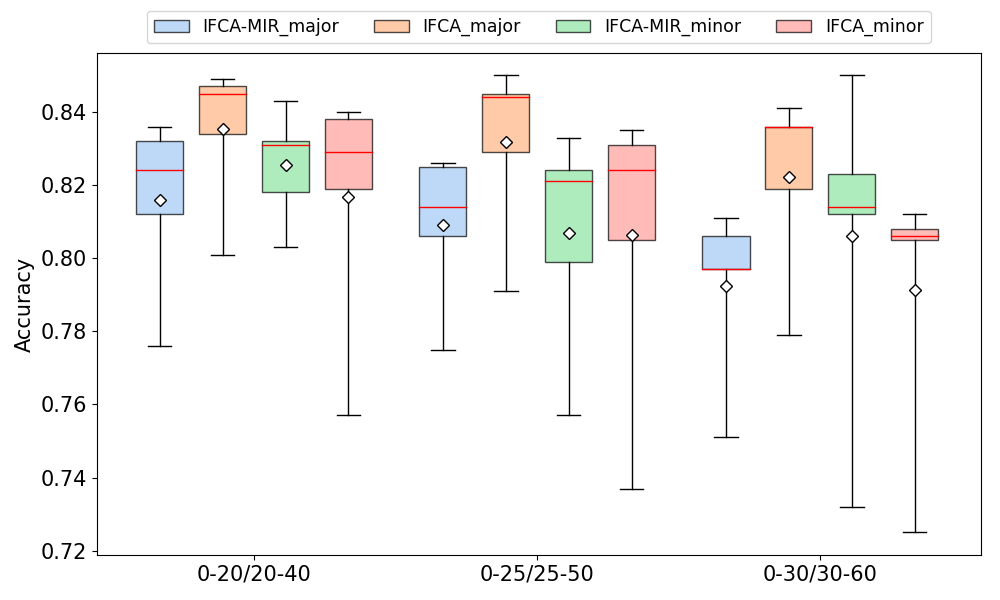}
    \caption{MNIST}
    \label{fig:mnist_dist_acc}
  \end{subfigure}
  \hspace{0.001\textwidth} 
  \begin{subfigure}[b]{0.5\textwidth}
    \centering
    \includegraphics[width=\textwidth]{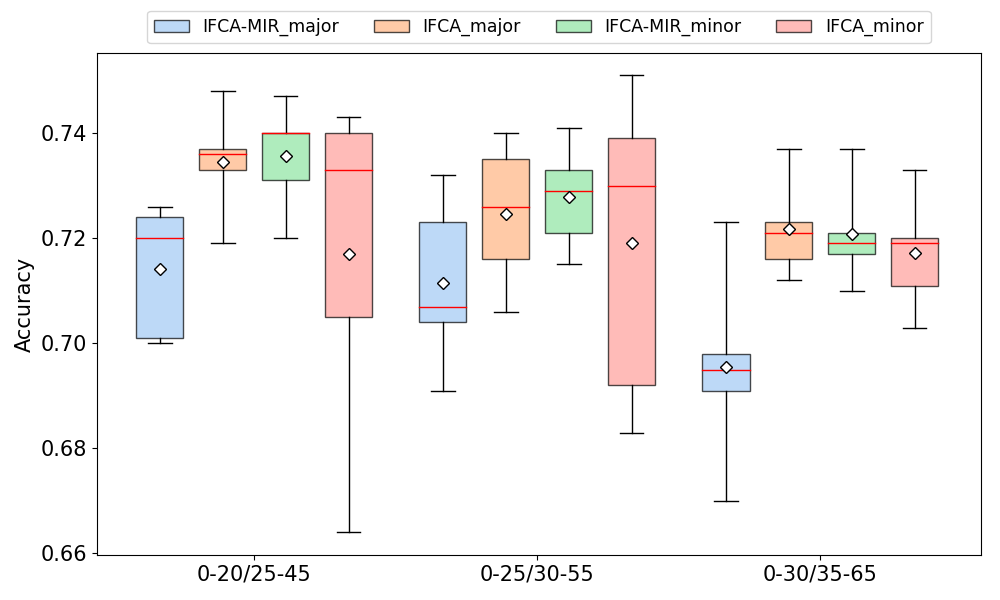}
    \caption{FEMNIST}
    \label{fig:femnist_dist_acc}
  \end{subfigure}
  \hspace{0.001\textwidth} 
  \begin{subfigure}[b]{0.5\textwidth}
    \centering
    \includegraphics[width=\textwidth]{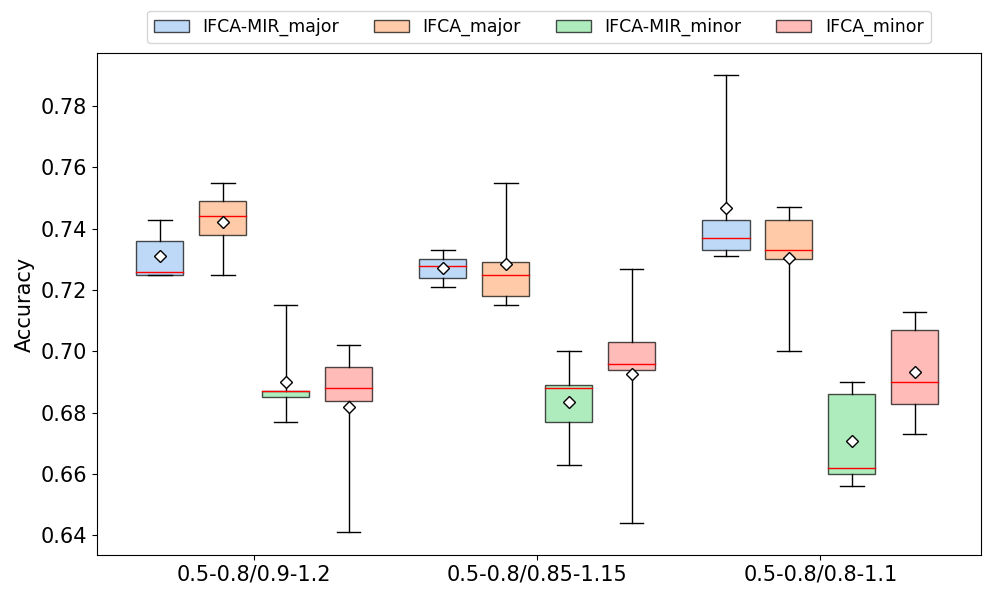}
    \caption{CIFAR-10}
    \label{fig:cifar10_dist_acc}
  \end{subfigure}
  \caption{Model accuracy results for each dataset with varying image deformation ranges}
  \label{fig:dist_acc}
\end{figure}

Figure \ref{fig:dist_acc} presents the results when the degree of image deformation was modified while keeping the minority client ratio fixed. Specifically, the minority ratio was fixed at 15\% for MNIST and FEMNIST, and at 20\% for CIFAR-10. The experiment was conducted by varying image rotation and brightness levels. The overall trend was consistent with the previous results. However, for CIFAR-10, the accuracy of the minority model under \sys decreased as the boundary between the majority and minority group distributions became less distinct. This is because, as the distributional differences between the two groups narrowed, more minority clients migrated to the majority group, reducing the amount of data available for training the minority model. Nevertheless, even in these cases, the decline in accuracy was minimal and remained comparable to that of the original IFCA. Additionally, as the range of image deformation increases, the training complexity rises, resulting in an overall decrease in accuracy.

\begin{table}
  \caption{Comparison of Average Model Accuracy Across Different Datasets}
  \label{tab:accuracy_comparison}
  \centering
  \begin{tabular}{lccc}
    \toprule
    Accuracy & MNIST & FEMNIST & CIFAR-10 \\
    \midrule
    IFCA \cite{ghosh} & 0.816 & 0.727 & 0.725 \\
    \sys & 0.797 & 0.704 & 0.719 \\
    \bottomrule
  \end{tabular}
\end{table}

Table \ref{tab:accuracy_comparison} presents the overall accuracy averages, which combine the accuracies of both the Major and Minority models. As shown in the tables, despite the differences between the Major and Minority models, the overall model accuracies of the original IFCA and \sys remain comparable.

\subsection{MIA vulnerability}

We conducted a comparative evaluation of MIA vulnerability between the original IFCA and \sys methods, using the same experimental setup as the model accuracy experiments.

\begin{table}
  \caption{Comparison of Average MIA accuracy Across Different Datasets}
  \label{tab:mia_comparison}
  \centering
  \begin{tabular}{ccccc}
    \toprule
     &  & MNIST & FEMNIST & CIFAR-10 \\
    \midrule
    \multirow{2}{*}{IFCA \cite{ghosh}} & Major & 0.521 & 0.537 & 0.604 \\
                              & Minor & 0.761 & 0.722 & 0.785 \\
    \midrule
    \multirow{2}{*}{\sys} & Major & 0.530 & 0.525 & 0.595 \\
                              & Minor & 0.791 & 0.733 & 0.792 \\
    \bottomrule
  \end{tabular}
\end{table}

Table \ref{tab:mia_comparison} presents the MIA accuracy of the Majority and Minority models for each dataset, comparing the original IFCA with \sys. If a client's MIA threshold is lower than this accuracy, it is considered a violation of MIA.

\begin{figure}[htbp]
  \centering
  \begin{subfigure}[b]{0.5\textwidth}
    \centering
    \includegraphics[width=\textwidth]{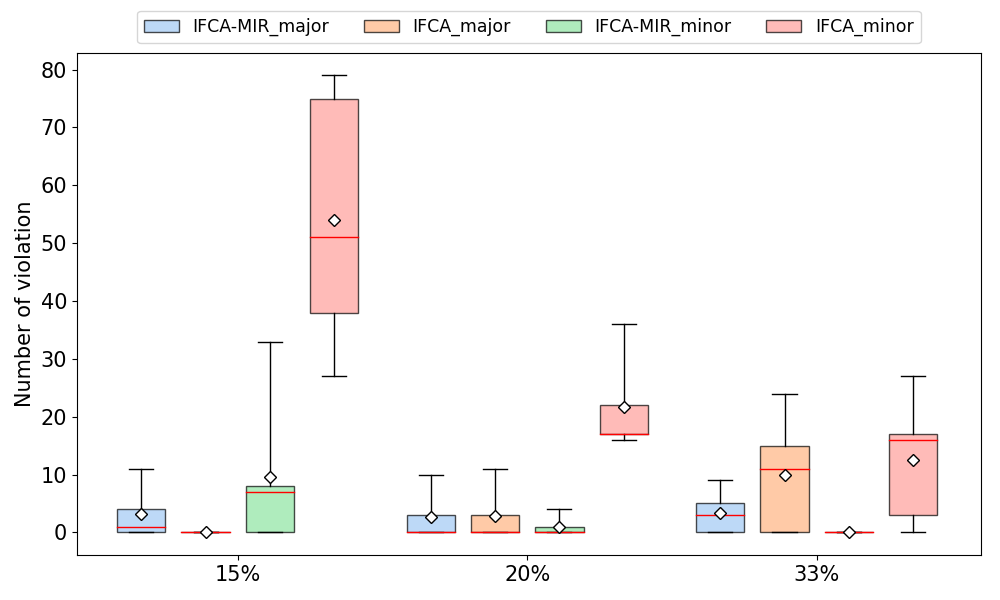}
    \caption{MNIST}
    \label{fig:mnist_num_mia}
  \end{subfigure}
  \hspace{0.001\textwidth} 
  \begin{subfigure}[b]{0.5\textwidth}
    \centering
    \includegraphics[width=\textwidth]{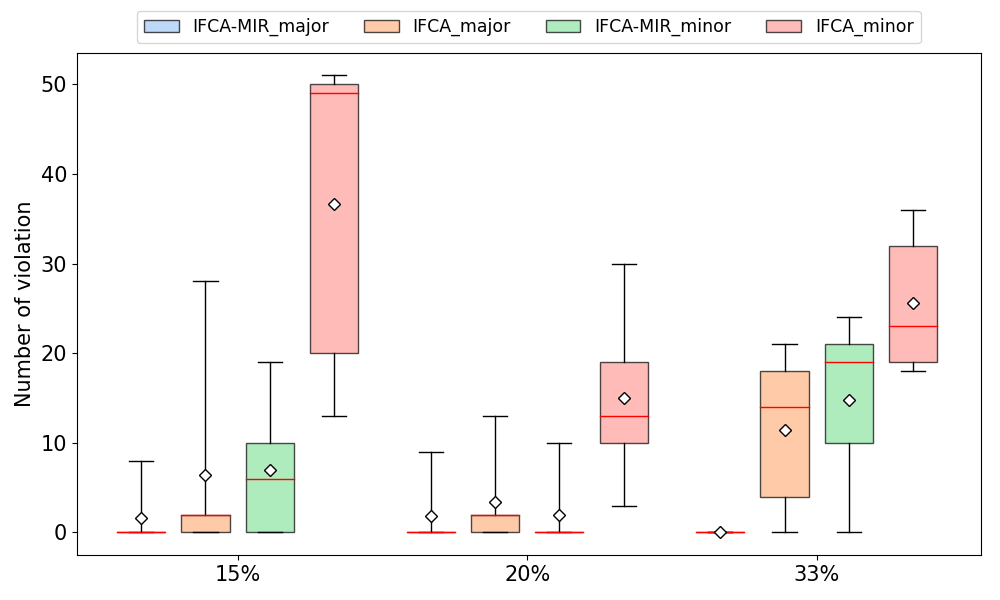}
    \caption{FEMNIST}
    \label{fig:femnist_num_mia}
  \end{subfigure}
  \hspace{0.001\textwidth} 
  \begin{subfigure}[b]{0.5\textwidth}
    \centering
    \includegraphics[width=\textwidth]{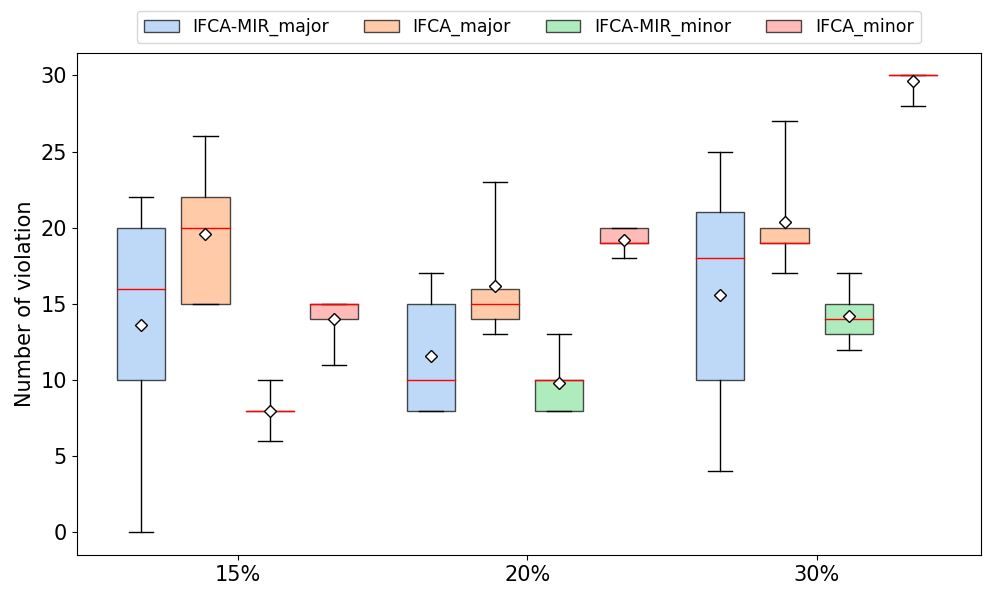}
    \caption{CIFAR-10}
    \label{fig:cifar10_num_mia}
  \end{subfigure}
  \caption{Number of MIA violation for each dataset with varying minority dataset sizes}
  \label{fig:num_mia}
\end{figure}

\begin{figure}[htbp]
  \centering
  \begin{subfigure}[b]{0.5\textwidth}
    \centering
    \includegraphics[width=\textwidth]{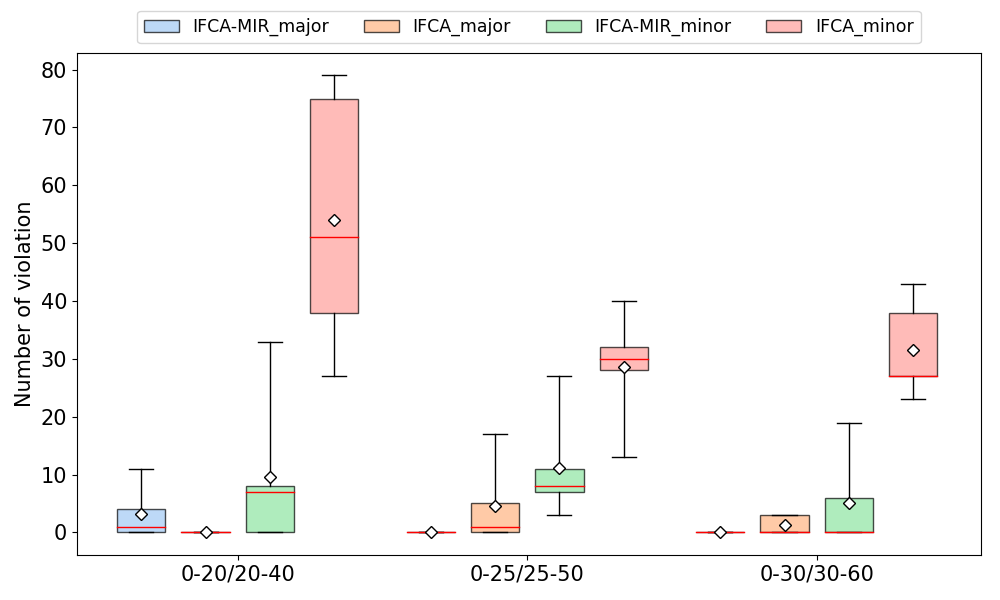}
    \caption{MNIST}
    \label{fig:mnist_dist_mia}
  \end{subfigure}
  \hspace{0.001\textwidth} 
  \begin{subfigure}[b]{0.5\textwidth}
    \centering
    \includegraphics[width=\textwidth]{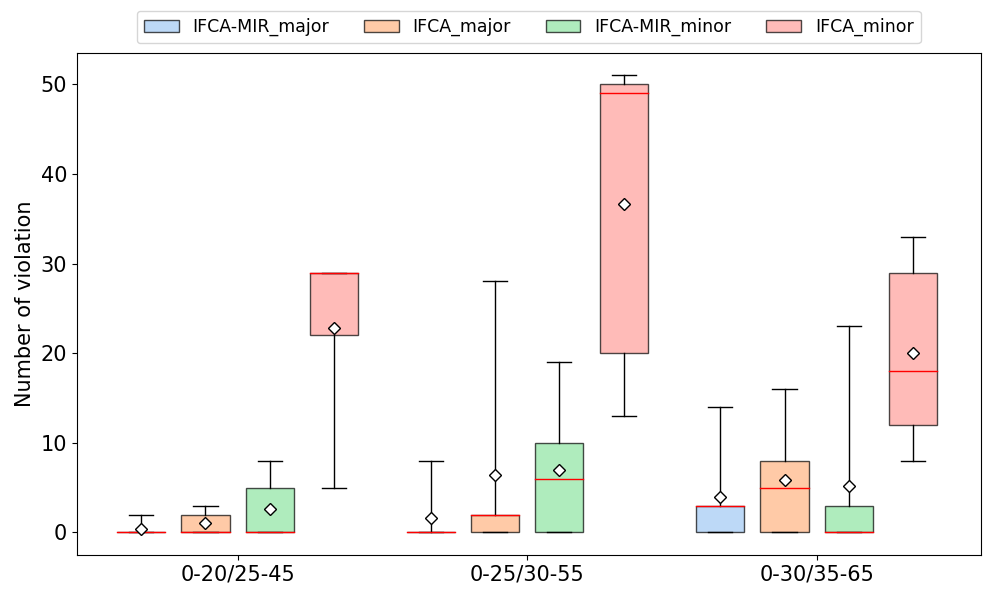}
    \caption{FEMNIST}
    \label{fig:femnist_dist_mia}
  \end{subfigure}
  \hspace{0.001\textwidth} 
  \begin{subfigure}[b]{0.5\textwidth}
    \centering
    \includegraphics[width=\textwidth]{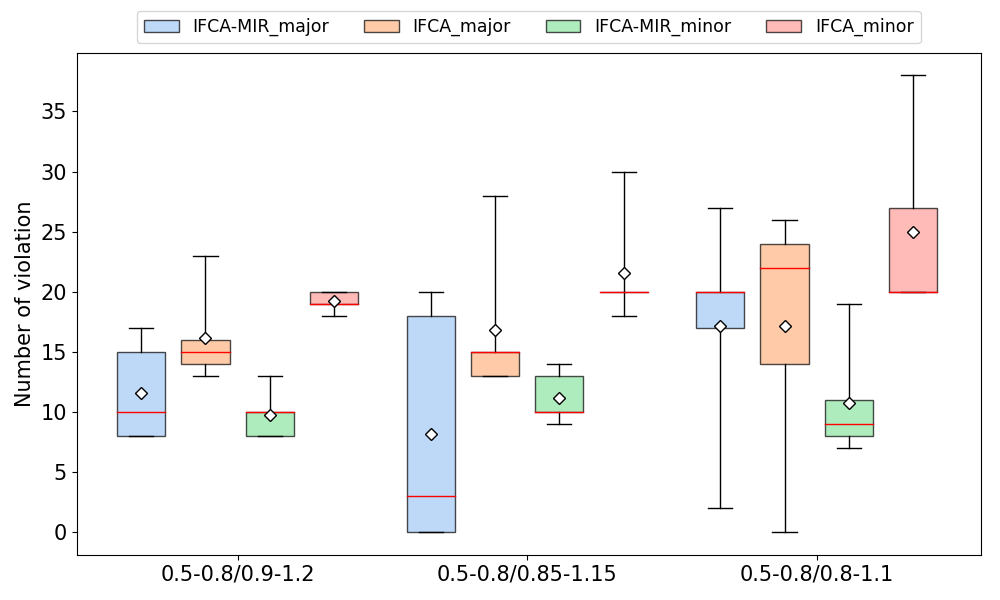}
    \caption{CIFAR-10}
    \label{fig:cifar10_dist_mia}
  \end{subfigure}
  \caption{Number of MIA violation for each dataset with varying image deformation ranges}
  \label{fig:dist_mia}
\end{figure}

Figures \ref{fig:num_mia} and \ref{fig:dist_mia} illustrate the results of this comparison. The experimental findings indicate that \sys significantly reduces MIA vulnerability compared to the original IFCA. This improvement can be attributed to the fact that the proposed method enables privacy-sensitive clients to select a safer group rather than solely optimizing for loss. Our experiments demonstrate that even without additional weight optimization, allowing clients to make group selections based on their privacy preferences leads to improved outcomes. This highlights the importance of integrating privacy considerations alongside traditional performance metrics, such as loss, in personalized model training. The results confirm that \sys effectively reduces the number of clients exposed to MIA risks.

By considering both loss and privacy concerns, the proposed method provides a flexible and adaptive framework that allows clients to balance performance and privacy based on their individual preferences.

\subsection{Fairness}
\begin{figure*}[htbp]
  \centering
  \begin{subfigure}[b]{0.9\textwidth}
    \centering
    \includegraphics[width=\textwidth]{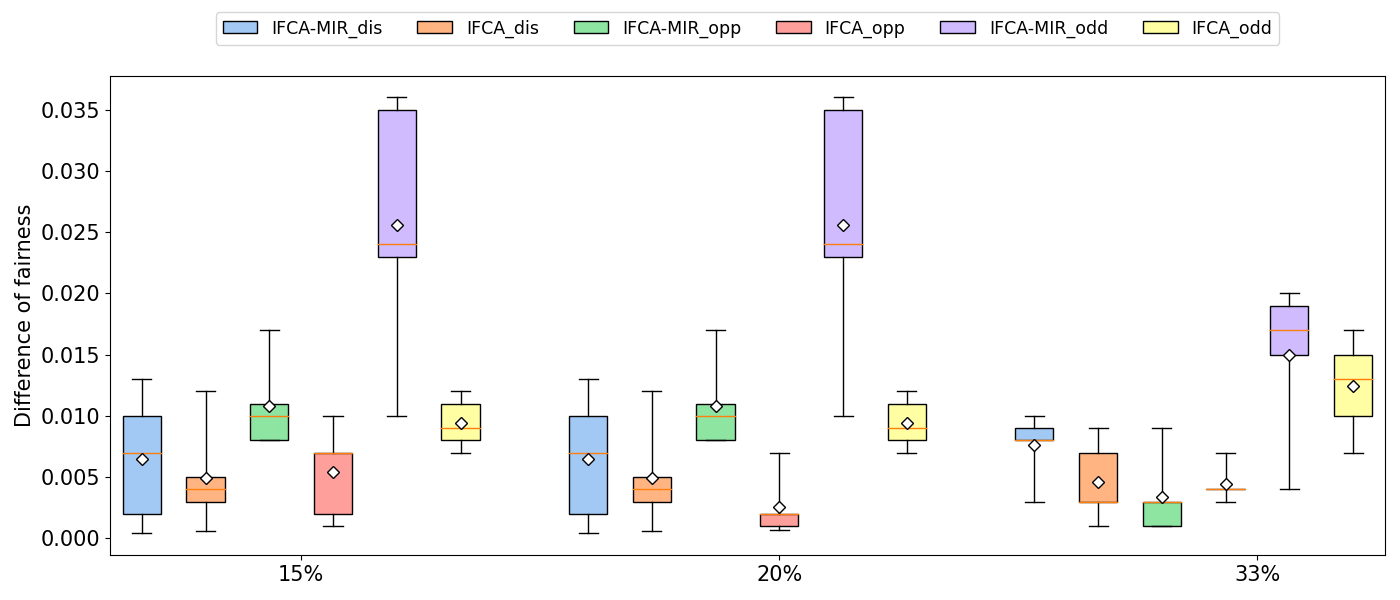}
    \caption{Varying minority dataset sizes}
    \label{fig:mnist_num_fairness}
  \end{subfigure}

  \begin{subfigure}[b]{0.9\textwidth}
    \centering
    \includegraphics[width=\textwidth]{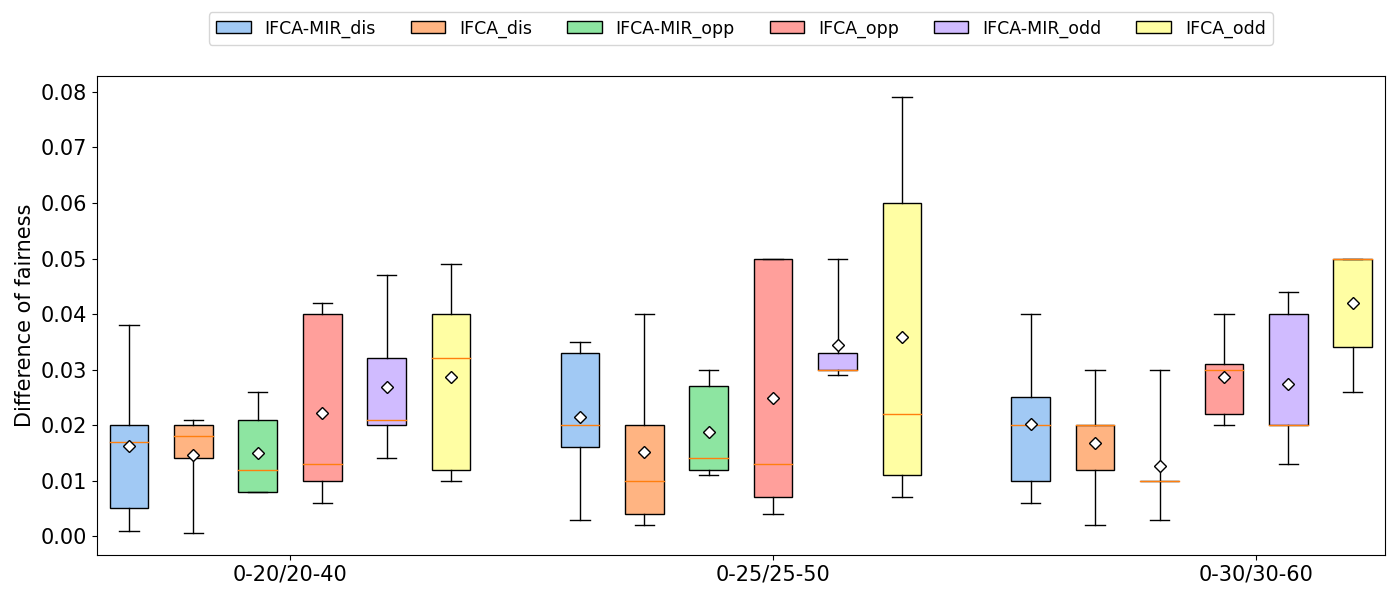}
    \caption{Varying image deformation ranges}
    \label{fig:mnist_dist_fairness}
  \end{subfigure}
  \caption{Fairness comparison for MNIST dataset with varying minority dataset sizes and image deformation ranges}
  \label{fig:mnist_fairness}
\end{figure*}

\begin{figure*}[htbp]
  \centering
  \begin{subfigure}[b]{0.9\textwidth}
    \centering
    \includegraphics[width=\textwidth]{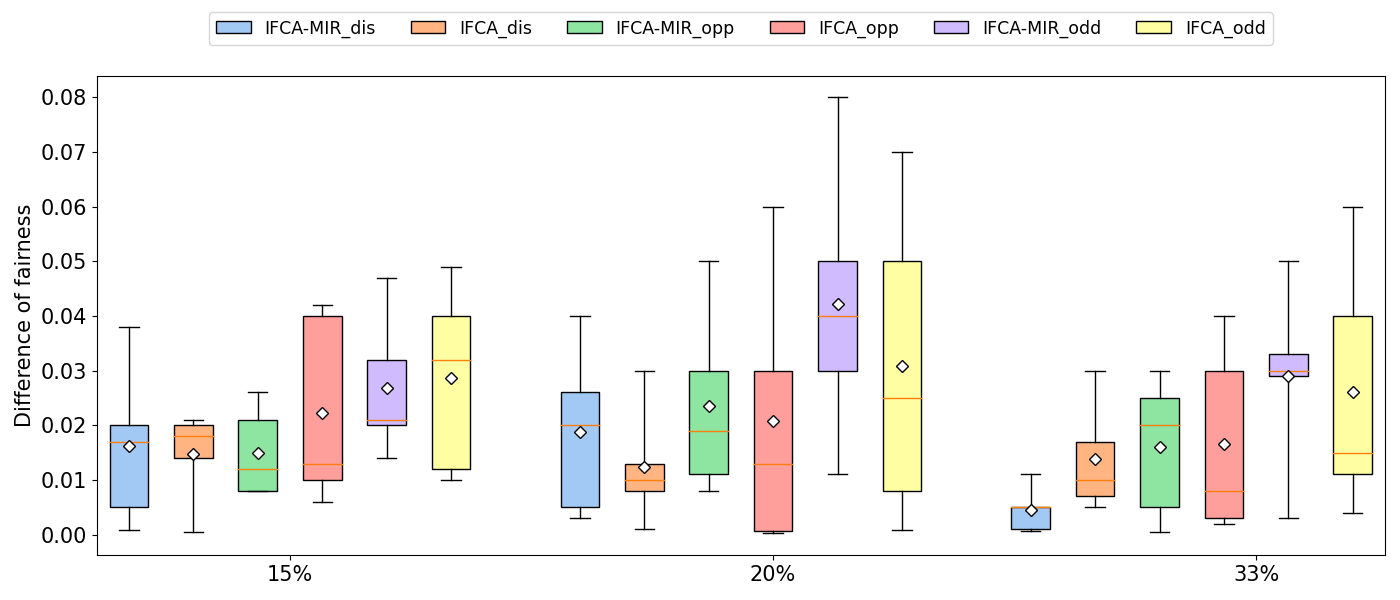}
    \caption{Varying minority dataset sizes}
    \label{fig:femnist_num_fairness}
  \end{subfigure}

  \begin{subfigure}[b]{0.9\textwidth}
    \centering
    \includegraphics[width=\textwidth]{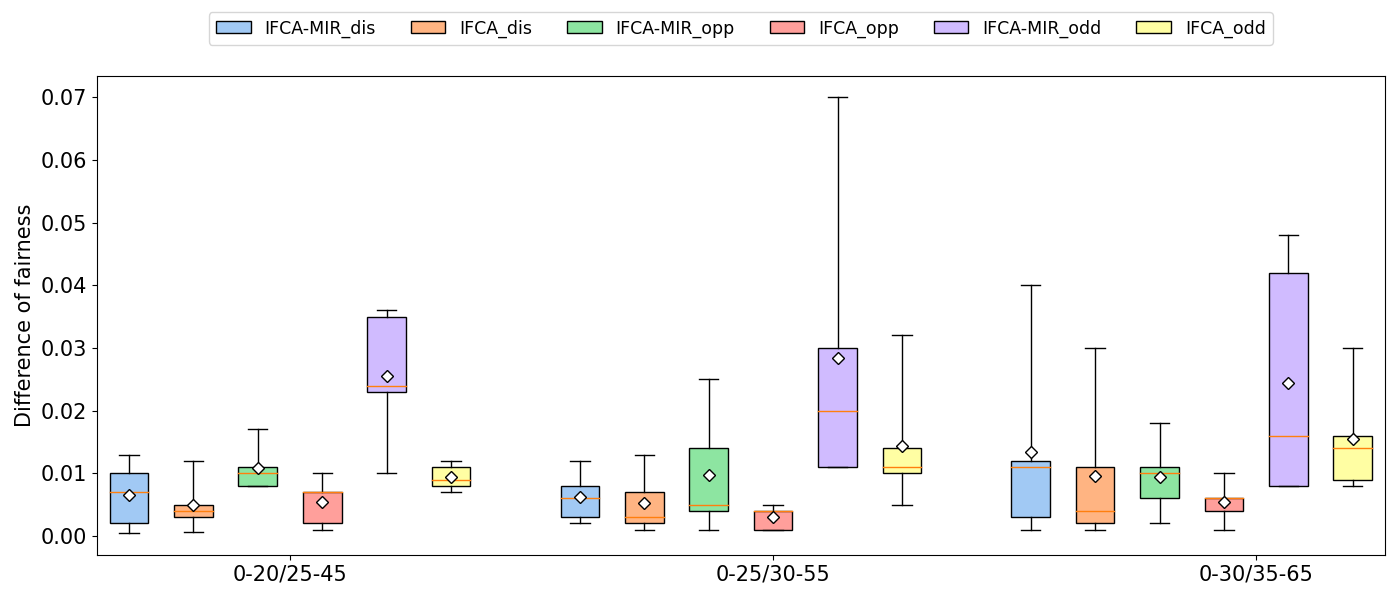}
    \caption{Varying image deformation ranges}
    \label{fig:femnist_dist_fairness}
  \end{subfigure}
    \caption{Fairness comparison for FEMNIST dataset with varying minority dataset sizes and image deformation ranges}
    \label{fig:femnist_fairness}
\end{figure*}

We conducted experiments to evaluate the fairness of the proposed method using the MNIST and FEMNIST datasets. As discussed earlier, PFL has been shown to enhance fairness compared to centralized learning methods \cite{galli2023advancing}. In this experiment, we aimed to assess the impact of the proposed \sys algorithm on fairness in comparison to the original IFCA approach.

The experimental results indicate that \sys does not introduce differences in fairness compared to the original IFCA method in Figure \ref{fig:mnist_fairness} and Figure \ref{fig:femnist_fairness}. Notably, on the MNIST dataset, \sys demonstrated slightly improved fairness over the original IFCA. The absolute fairness values for both methods remained close to zero, suggesting that both the original IFCA and \sys contribute to fairer outcomes in federated learning environments.

Overall, the findings demonstrate that \sys preserves the fairness advantages of personalized federated learning without introducing substantial trade-offs compared to the original IFCA. 

\subsection{Convergence}
\begin{figure}[htbp]
\centerline{\includegraphics[width=0.5\textwidth]{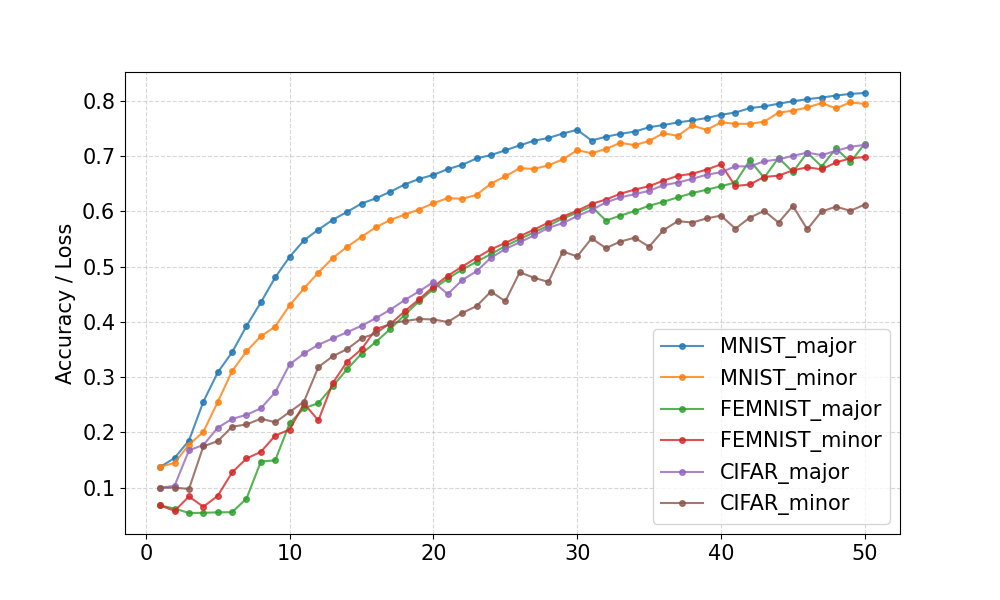}}
\caption{Convergence graph of majority and minority models for each dataset}
\label{fig:convergence}
\end{figure}

We demonstrate the theoretical convergence of our proposed technique in Section \ref{subsec:convergence} and validate it experimentally. As shown in Figure \ref{fig:convergence}, the models for both the majority and minority groups trained using \sys successfully converge across all datasets.

\subsection{Discussion}\label{subsec:discussion}
Our proposed \sys method successfully reduces MIA accuracy while maintaining overall model performance and fairness. However, during this balancing process, we observed an shift of minority group clients to the majority group. This phenomenon arises because the majority group generally offers better privacy protection compared to the minority group. When the accuracy difference between the majority and minority groups is small, privacy-sensitive clients accept a slight accuracy loss to migrate to the majority group, as it provides lower MIA risk.

While this trade-off is a fundamental aspect of our method, it also introduces a potential over-concentration in the majority group, which can significantly degrade the performance of the minority group. As shown in Figure \ref{fig:cifar10_num_acc}, such shifts may reduce the accuracy of the minority group, and it might be made an unacceptable level. From a global optimization perspective, the server may need additional mechanisms to prevent excessive client migration to the majority group. One possible solution is to introduce an incentive-based balancing mechanism, where privacy-sensitive clients in the minority group are offered rewards to maintain diversity within clusters.

Another key consideration in our approach is the reliance on the server to act as a red team for MIA evaluation. Our method assumes that the server performs MIA, evaluates its risk, and provides this information to clients, enabling privacy-aware model selection. However, this approach is only valid under the assumption that the server is fully trusted. In settings where the server cannot be trusted, clients may instead perform MIA evaluation locally to ensure privacy-preserving model selection. 

In such cases, clients could perform MIA without training additional shadow models, relying only on their local models and the models received from the server. While this approach eliminates server trust dependency, it also introduces computational overhead for individual clients. Nonetheless, in cross-silo federated learning environments, where clients typically have higher computational capacity, such an approach may be feasible.

\section{Conclusion}
In this paper, we proposed \sys, an enhanced clustering-based personalized federated learning algorithm that mitigates Membership Inference Attack (MIA) vulnerability while maintaining model accuracy and fairness. Unlike the original IFCA, which selects models based solely on empirical loss, \sys incorporates MIA risk assessment into the model selection process. This enables clients to balance privacy protection and model performance based on their individual sensitivity to MIA.

Through extensive experiments on the MNIST, FEMNIST, and CIFAR-10 datasets, we demonstrated that \sys effectively reduces the number of clients exposed to MIA risks compared to the original IFCA. Our findings show that while majority group accuracy remains comparable between both methods, minority group accuracy is either maintained or improved, as privacy-sensitive clients migrate to safer clusters. Moreover, our fairness evaluation confirmed that \sys preserves the fairness benefits of personalized federated learning without introducing significant trade-offs. Overall, \sys provides a practical and adaptive approach to mitigating MIA risks in federated learning while preserving model accuracy and fairness.

Our future research direction is to develop an incentive mechanism to prevent excessive migration of minority group clients to the majority group. To achieve this, the server should act as a coordinator that seeks an Pareto optimality between overall model accuracy and MIA vulnerability, providing appropriate incentives to clients. Another promising research direction is the application of DP. DP is a well-established technique for mitigating MIA risks, whereas our study focused on reducing MIA vulnerability without relying on DP. Introducing DP may lead to a trade-off between privacy protection and model accuracy. Therefore, exploring optimal DP strategies that minimize accuracy degradation while ensuring robustness against MIA would be a valuable avenue for future research.

\begin{acks}
To Robert, for the bagels and explaining CMYK and color spaces.
\end{acks}


\bibliographystyle{plain}
\bibliography{references}

\appendix
\section{Appendices}

\subsection{Model and parameter}

The CNN model for MNIST datasets is as follows:
\begin{itemize}
\item First Convolutional Layer: Takes an input of shape (1, 28, 28) and applies a convolution with 32 filters of size 3x3, followed by ReLU activation and 2x2 max-pooling. This layer outputs a feature map of shape (32, 14, 14).
\item Second Convolutional Layer: This layer takes the (32, 14, 14) input and applies 64 filters of size 3x3, followed by ReLU activation and 2x2 max-pooling. The output is a feature map of shape (64, 7, 7).
\item Fully Connected Layer: Flattens the feature maps and connects them to a fully connected layer with 10 output neurons, representing the number of classes. Xavier initialization is used for the weights in this layer.
\end{itemize}

The CNN model for FEMNIST datasets is as follows:
\begin{itemize}
\item First Convolutional Layer: Applies 32 filters of size 3x3 to the input, producing feature maps. The ReLU activation function is used after the convolution.
\item Second Convolutional Layer: Takes the output from the first layer and applies 64 filters of size 3x3, followed by ReLU activation.
\item Max-Pooling Layer: Reduces the spatial dimensions of the feature maps using a 2x2 max-pooling operation.
\item Fully Connected Layers:
The first fully connected layer flattens the feature maps and connects them to 128 neurons, followed by ReLU activation.
The second fully connected layer maps these 128 neurons to 62 output features, representing the final classification layer.

\end{itemize}

The specific model structure for CIFAR-10 datasets is as follows:

\begin{itemize}
\item Convolutional Layers: The convolutional layers use 3x3 kernels and varying numbers of filters as specified by the selected VGG configuration . The network progressively increases the number of filters (e.g., 64, 128, 256, 512) and uses ReLU activation functions. Batch normalization is applied after each convolutional layer to stabilize and accelerate training.
\item Pooling Layers: Max-pooling layers are used after specific sets of convolutional layers to downsample the feature maps, reducing their spatial dimensions.
\item Fully Connected Layers:
After the convolutional layers, the output is flattened and passed through three fully connected layers.
The first two fully connected layers have 512 neurons each with ReLU activation.
The final fully connected layer maps the features to 10 output classes.
A log-softmax function is applied to the output to produce log-probabilities for classification.

\end{itemize}

\subsection{Image sample}
\begin{figure}[htbp]
    \centering
    \begin{subfigure}{0.45\textwidth}
        \centering
        \includegraphics[width=\textwidth]{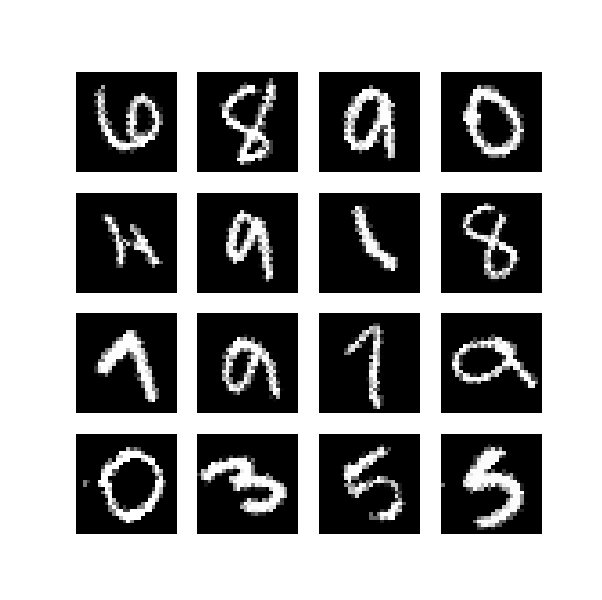}
        \caption{Majority samples}
        \label{fig:subfig1}
    \end{subfigure}
    \begin{subfigure}{0.45\textwidth}
        \centering
        \includegraphics[width=\textwidth]{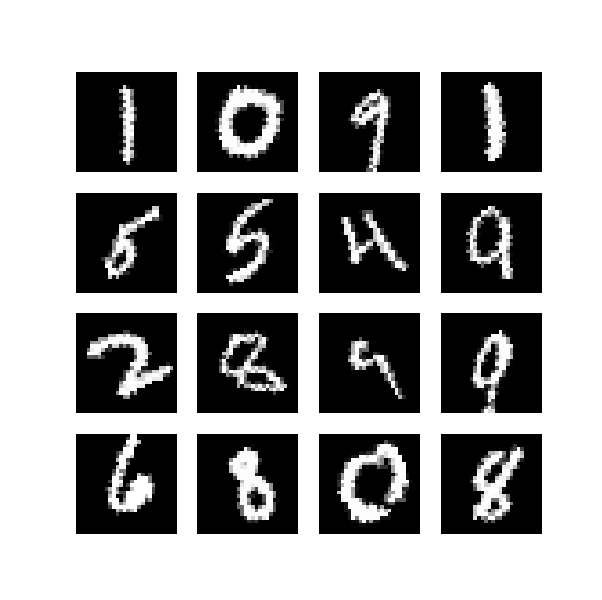}
        \caption{Minority samples}
        \label{fig:subfig2}
    \end{subfigure}
    \caption{Data samples for majority and minority in MNIST dataset}
    \label{fig:main}
\end{figure}

\begin{figure}[htbp]
    \centering
    \begin{subfigure}{0.45\textwidth}
        \centering
        \includegraphics[width=\textwidth]{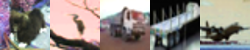}
        \caption{Majority samples}
        \label{fig:subfig3}
    \end{subfigure}
    \begin{subfigure}{0.45\textwidth}
        \centering
        \includegraphics[width=\textwidth]{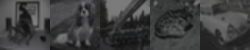}
        \caption{Minority samples}
        \label{fig:subfig4}
    \end{subfigure}
    \caption{Data samples for majority and minority in CIFAR-10 dataset}
    \label{fig:sample}
\end{figure}

\end{document}